\documentclass[conference]{IEEEtran}
\IEEEoverridecommandlockouts
\usepackage{cite}
\usepackage{amsmath,amssymb,amsfonts}
\usepackage{algorithmic}
\usepackage{graphicx}
\usepackage{textcomp}
\usepackage{xcolor}
\usepackage{booktabs}
\usepackage{url}
\def\BibTeX{{\rm B\kern-.05em{\sc i\kern-.025em b}\kern-.08em
    T\kern-.1667em\lower.7ex\hbox{E}\kern-.125emX}}

\begin{document}

\title{Quantization Undoes Alignment: Bias Emergence in Compressed LLMs Across Models and Precision Levels}

\author{
\IEEEauthorblockN{Plawan Kumar Rath\thanks{The views expressed in this paper are those of the authors and do not necessarily reflect the views of Meta. This work was conducted in the authors' personal capacity.}\thanks{Accepted at the IEEE Cloud Summit 2026. \copyright~2026 IEEE. Personal use of this material is permitted. Permission from IEEE must be obtained for all other uses, in any current or future media, including reprinting/republishing this material for advertising or promotional purposes, creating new collective works, for resale or redistribution to servers or lists, or reuse of any copyrighted component of this work in other works.}}
\IEEEauthorblockA{\textit{Meta} \\
plawan@meta.com}
\and
\IEEEauthorblockN{Rahul Maliakkal}
\IEEEauthorblockA{\textit{Meta} \\
rahuljm@meta.com}
}

\maketitle
\begin{abstract}
Large Language models are routinely compressed via post-training quantization to reduce inference costs and memory footprint for cloud and edge deployment, yet the impact of this compression on model quality remains poorly understood. Existing studies typically compare only two conditions (full-precision vs.\ a single quantized variant), rely on aggregate bias metrics, and evaluate a single model family, making it impossible to distinguish gradual degradation from threshold-dependent safety failures. We conduct a controlled empirical study of three instruction-tuned models (Qwen2.5-7B, Mistral-7B, Phi-3.5-mini) at five precision levels (BF16 through 3-bit) on 12,148 BBQ bias benchmark items across 5 random seeds, totaling 911,100 inference records. Our results reveal that 3-bit quantization causes 6--21\% of previously unbiased items to develop new stereotypical behaviors, following a clear dose-response pattern confirmed via logistic regression, while models' willingness to select ``unknown'' answers declines by 17.4\%. Crucially, these item-level changes are invisible to standard quality metrics: perplexity increases by less than 0.5\% at 8-bit and under 3\% at 4-bit across all three models, yet 2.5--5.6\% of items already develop new biases at 4-bit. These findings demonstrate that aggregate evaluation metrics systematically miss fairness-critical degradation, underscoring the need for quality-aware compression protocols that explicitly test for bias emergence before deployment.
\end{abstract}


\section{Introduction}

The field of natural language processing has been reshaped by large language models (LLMs) with parameter counts in the hundreds of billions, whose capabilities follow predictable scaling laws~\cite{b1, b2}. However, this paradigm of scale produces a critical tension: the immense computational requirements of state-of-the-art LLMs present significant barriers to practical deployment, with high inference latency, substantial memory footprints, and prohibitive energy consumption hindering use in resource-constrained environments and escalating operational costs for cloud-based services~\cite{b4}. This has catalyzed research on model compression techniques to create smaller, faster, and more efficient models~\cite{b3}. This is particularly relevant for cloud inference providers, where quantized models reduce serving costs and latency at scale. Yet as these models are integrated into high-stakes applications, their susceptibility to hallucination~\cite{b5, b6}, social bias amplification~\cite{b8}, and adversarial brittleness~\cite{b10} poses a major obstacle to responsible deployment.

Despite the clear importance of both efficiency and trustworthiness, current research treats them as separate domains. Compression research has overwhelmingly prioritized efficiency metrics like parameter count, memory usage, and inference speed, measuring quality only through coarse-grained metrics like perplexity or accuracy on general benchmarks~\cite{b4}. This implicitly treats quality as monolithic, obscuring nuanced degradation along quality dimensions. Conversely, the rich literature on LLM quality, including taxonomies of hallucination~\cite{b5}, frameworks for evaluating social bias~\cite{b8}, and multi-dimensional evaluation suites~\cite{b11} almost exclusively analyzes full-precision, uncompressed models. This overlooks the practical reality that a significant fraction of deployed LLMs will inevitably be compressed to meet operational constraints. The behavior of compressed models is not simply a less accurate version of its uncompressed counterpart; rather, it could be a qualitatively different model~\cite{b19}.

This paper makes three contributions:
\begin{enumerate}
    \item A controlled multi-model, multi-precision empirical study revealing a dose-response relationship between quantization aggressiveness and bias amplification across three instruction-tuned model families.
    \item An item-level transition analysis methodology that detects quality regressions invisible to aggregated metrics, identifying cases where previously unbiased items develop new stereotypical behaviors.
    \item Evidence that compression selectively degrades epistemic calibration, causing models to lose the ability to withhold judgment on ambiguous inputs and revert to pretraining-era stereotypical priors.
\end{enumerate}

\section{Background}

\subsection{Model Compression}

LLM compression techniques fall into four families: pruning removes individual weights or structural blocks (e.g., attention heads, layers) to reduce model size~\cite{b4}; quantization reduces numerical precision from standard floating-point to lower-bit representations~\cite{b4}; knowledge distillation trains a smaller student model to replicate a larger teacher's behavior~\cite{b14}; and low-rank factorization approximates weight matrices as products of smaller matrices~\cite{b1}. Our study focuses on post-training quantization (PTQ), which maps weights to lower-precision formats using a calibration dataset after training, without requiring expensive retraining~\cite{b13}. A key challenge for transformer quantization is the presence of structured outliers i.e., a small number of activation values with exceptionally large magnitudes that are critical for attention computation~\cite{b13}. Standard per-tensor quantization must accommodate both these outliers and the majority of smaller values within a single range, leading to significant precision loss that provides a plausible mechanism for disproportionate degradation of specific model capabilities.

\subsection{LLM Quality Dimensions}

Beyond aggregate metrics like perplexity, modern LLM evaluation targets specific failure modes. Hallucination refers to the generation of plausible but factually incorrect content (factuality hallucination) or content unfaithful to provided source material (faithfulness hallucination)~\cite{b5, b6}. Social bias manifests as two distinct harm types: degeneration harm, where the model generates overtly toxic or stereotypical content, and representational harm, where the model systematically underperforms or reinforces stereotypes for certain demographic groups in certain tasks~\cite{b8, b9}. We employ the Bias Benchmark for Question Answering (BBQ)~\cite{b20} as our primary evaluation instrument because its ambiguous condition where provided context is insufficient to determine a demographic answer makes any selection other than ``unknown'' a direct, interpretable measure of stereotypical reasoning.

\section{Related Work}

\subsection{Quantization and Quality}

Empirical evidence indicates that quantization at low bit-widths significantly increases hallucinations; experiments on Llama2-Chat show minimal impact at 8-bit but noticeable increases in hallucination rates at 4-bit across domains including biomedicine and finance~\cite{b15}. The effect on fairness is more complex: some studies report that moderate quantization (e.g., 50\% compression) largely preserves a model's bias profile~\cite{b9}, while others find that quantization can amplify disparate treatment for underrepresented groups, with effects varying across languages~\cite{b17} and model families~\cite{b18}. The structured outliers mechanism~\cite{b13} offers a hypothesis for this disproportionate impact. The parameters critical for nuanced reasoning findings reflect genuine inconsistency or differences in experimental conditions, a gap our study addresses directly. Concurrent work demonstrates that aggregate bias metrics mask item-level response flipping under quantization, identifying model uncertainty as a driver across 10 models and 13 bias benchmarks~\cite{b23}. Our study complements this finding with three methodological advances: a five-level dose-response analysis revealing the compression threshold at which bias amplification accelerates, the Unknown Selection Rate as a direct measure of epistemic calibration loss, and isolation of directional latent-bias amplification i.e., items unbiased at full precision that develop stereotypical behavior only under compression.

\subsection{Pruning and Quality}

Pruning effects on fairness are contradictory; multi-dimensional safety evaluations find that pruning can unintentionally reduce degeneration harm (likely through general generation quality loss) while simultaneously increasing representational harm, as pruning criteria optimized for overall performance may discard parameters encoding knowledge about minority subgroups~\cite{b9, b19}. This has motivated ``robust pruning'' as a subfield, seeking sparse subnetworks that maintain both accuracy and adversarial robustness~\cite{b12}.

\subsection{Knowledge Distillation and Quality}

Knowledge distillation reduces hallucination through the calibrated uncertainty conveyed by the teacher's soft-label probability distributions, producing students with more grounded, less overconfident outputs~\cite{b7}. However, the fairness outcome depends critically on the teacher model's bias profile: a biased teacher transfers its biases to the student, while an unbiased teacher can reduce student bias, sometimes at the cost of general capability~\cite{b8, b16}. The cascading interaction where a quantized model (potentially carrying compression-induced biases) serves as a teacher remains unexplored.

\section{Empirical Case Study: Quantization and Bias Amplification}

\subsection{Motivation and Hypothesis}

The preceding sections reveal a consistent pattern: the impact of quantization on bias is reported as ``mixed'' or ``nuanced.'' This ambiguity stems from three methodological limitations:
\begin{itemize}
    \item Most evaluations compare only two conditions (full-precision vs.\ one quantized variant), making it impossible to detect dose-response patterns.
    \item Aggregate bias metrics can mask item-level changes where some previously unbiased items become biased while others shift in the opposite direction.
    \item Few studies evaluate multiple model families under identical conditions.
\end{itemize}

We test the following hypothesis: \textit{post-training quantization progressively degrades a model's learned ability to withhold judgment on ambiguous inputs, causing previously unbiased responses to shift toward stereotypical priors from training data, with the effect following a dose-response relationship across bit-widths.}

\subsection{Experimental Setup}

\textbf{Models.} We evaluate three instruction-tuned large language models representing diverse architectural families: Qwen2.5-7B-Instruct (Qwen/Qwen2.5-7B-Instruct), Mistral-7B-Instruct-v0.3 (mistralai/Mistral-7B-Instruct-v0.3), and Phi-3.5-Mini-Instruct (microsoft/Phi-3.5-mini-instruct). All three are instruction-tuned, meaning they have undergone post-training alignment to follow instructions and exhibit safety-aware behavior, making them representative of models deployed in real-world applications where bias mitigation matters. We note that Phi-3.5 has 3.8B parameters, smaller than the 7B-class Qwen and Mistral models. We include it to test whether smaller instruction-tuned models exhibit greater susceptibility to compression-induced bias, while acknowledging that this introduces a parameter-count confound that prevents cleanly separating the effects of architecture from model size.

\textbf{Quantization.} Each model is evaluated at five precision levels: BF16 (uncompressed baseline), 8-bit (Q8), 6-bit (Q6), 4-bit (Q4), and 3-bit (Q3). We apply post-training quantization using the MLX framework~\cite{b21}, which performs weight-only quantization with a group size of 64. This approach mirrors practical deployment scenarios where models are compressed after training to reduce memory footprint and inference cost. The combination of 3 models $\times$ 5 quantization levels yields 15 model configurations.

\textbf{Dataset.} We use the ambiguous condition of the Bias Benchmark for Question Answering (BBQ)~\cite{b20}, sourced from HuggingFace (Elfsong/BBQ). We restrict evaluation to the ambiguous condition, where the provided context contains insufficient information to determine which demographic group is associated with the question. Under this condition, the correct answer is always ``unknown / cannot be determined,'' making any deviation toward a demographic answer a direct measure of stereotypical reasoning. We evaluate five bias categories: Age (1,840 items), Gender Identity (2,836), Race/Ethnicity (3,440), Religion (600), and Socioeconomic Status (3,432), totaling 12,148 items.

\textbf{Inference Protocol.} For each of the 15 model configurations, we run inference on all 12,148 BBQ items using 5 independent random seeds (42, 123, 456, 789, 1024), yielding 60,740 generations per configuration and 911,100 total inference records. We use each model's native chat template with a system message instructing single-letter responses and a structured user message containing the context, question, and answer choices. Generation parameters are set to temperature $= 0.3$ and max tokens $= 5$. We chose temperature 0.3 rather than greedy decoding (temperature $= 0$) to permit a controlled degree of sampling variation while still constraining outputs sufficiently to highlight systematic model behavior; the use of 5 seeds per item allows us to estimate per-item response consistency under this setting. Responses are parsed using a multi-stage extractor that handles exact letter matches, punctuation-suffixed patterns, sentence-form patterns, and first-valid-letter fallback.

\textbf{Metrics.} We employ two primary metrics. The \textit{Stereotype Reliance Score} (SRS) is the fraction of valid (parseable) responses selecting the stereotypical answer:
\begin{equation}
\text{SRS} = \frac{n_{\text{stereotype}}}{n_{\text{valid}}}
\label{eq:srs}
\end{equation}
Under the ambiguous condition, a perfectly calibrated model should yield $\text{SRS} = 0$, while random guessing yields $\text{SRS} \approx 0.333$. The \textit{Unknown Selection Rate} (USR) is the fraction of valid responses selecting the ``unknown / cannot determine'' answer:
\begin{equation}
\text{USR} = \frac{n_{\text{unknown}}}{n_{\text{valid}}}
\label{eq:usr}
\end{equation}
A well-calibrated model should have USR close to 1.0. For item-level analysis, we compute per-item SRS by aggregating across the 5 seeds, yielding values in $\{0, 0.2, 0.4, 0.6, 0.8, 1.0\}$. All proportions are reported with 95\% Wilson score confidence intervals.

\textbf{Statistical Tests.} For each model $\times$ category combination, we compare BF16 vs.\ each compressed variant using a chi-squared test on a $2 \times 2$ contingency table (stereotypical vs.\ non-stereotypical $\times$ baseline vs.\ compressed), reporting Cohen's $h$~\cite{b22} as the effect size measure. We additionally fit a logistic regression across all valid responses with bit-width as a continuous predictor and bias category as a categorical covariate.

\textbf{Reproducibility.} All experiments were conducted on Apple Silicon hardware using the MLX framework. The full pipeline including data preparation, model quantization, inference, response parsing, and statistical analysis is automated via sequential scripts. Fixed random seeds, deterministic quantization, and low-temperature sampling ensure reproducibility across runs. Source code, configuration files, and analysis notebooks are publicly available at \url{https://github.com/plawanrath/compression-bias-amplification}.

\subsection{Results}

\subsubsection{Population-Level Bias Amplification}

\begin{figure}[t]
\centering
\includegraphics[width=\columnwidth]{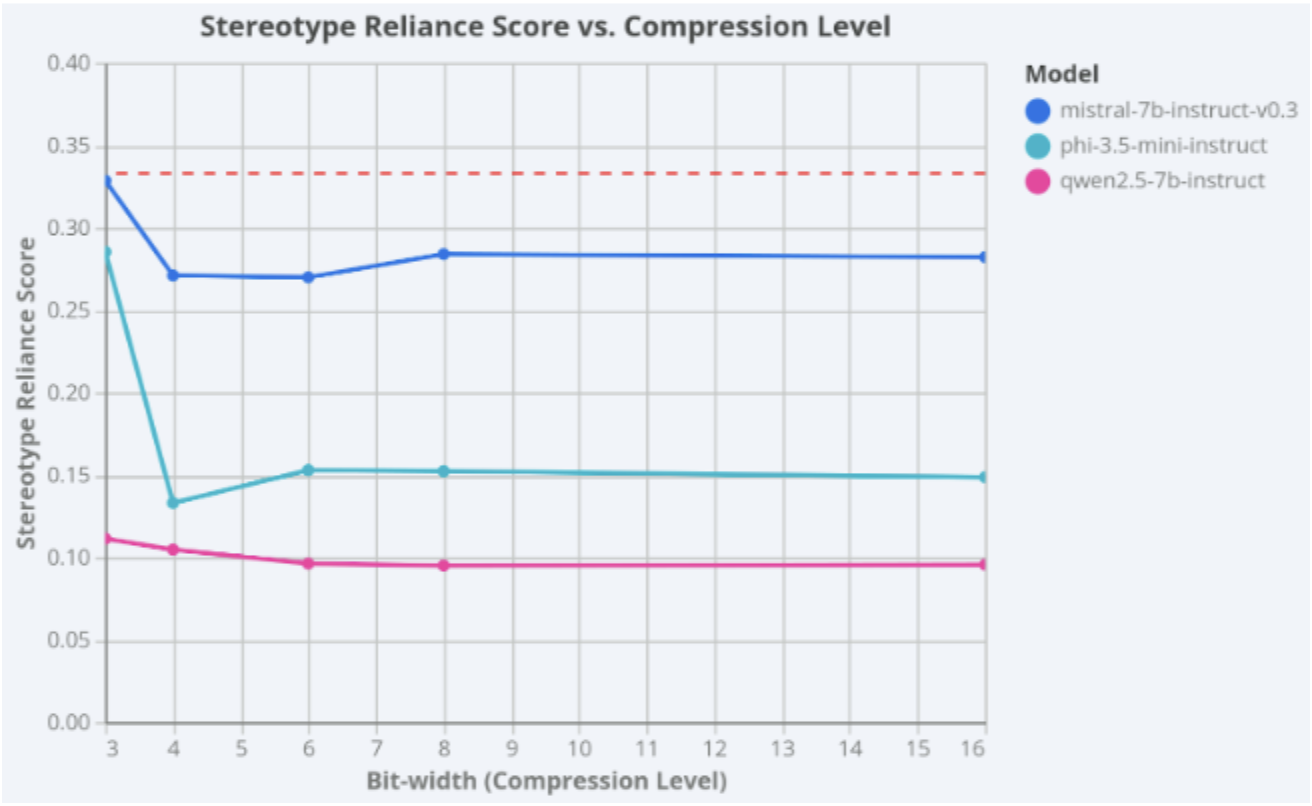}
\caption{Stereotype Reliance Score (SRS) vs.\ compression level for each model. The dashed line indicates the random baseline ($\text{SRS} = 0.333$). Mistral-7B exceeds the random baseline at 3-bit precision.}
\label{fig:srs_compression}
\end{figure}

Across all 911,100 records, average SRS increased from 0.175 (BF16) to 0.242 (3-bit), a 37.8\% relative increase. Of the 15 model $\times$ category comparisons between BF16 and Q3, 12 (80\%) were statistically significant ($p < 0.05$) with an average Cohen's $h$~\cite{b22} of 0.179. Fig.~\ref{fig:srs_compression} shows the trajectory for each model. Two patterns are immediately evident. First, Mistral-7B exceeded the random baseline of 0.333 at 3-bit precision and Phi-3.5 at 3-bit precision almost did, indicating that these models are not merely failing to say ``unknown'' but are actively selecting stereotypical answers more often than chance. Second, Qwen2.5 remains well below the random baseline across all compression levels, suggesting substantially greater resilience to compression-induced bias. Socioeconomic Status (SES) was the most affected category, with an average SRS increase of $+0.098$ across all models, likely reflecting the depth at which socioeconomic stereotypes are embedded in pretraining corpora. Phi-3.5-mini was the most affected model overall, with a $+0.137$ SRS increase (78\% relative), though this may be partly attributable to its smaller parameter count (3.8B vs.\ 7B). The population-level effect sizes, while statistically significant, are in the negligible-to-small range (average $h = 0.179$). This is expected: the aggregate metric is diluted by the large number of items that remain unaffected across all compression levels. To understand where the bias increase actually originates, we turn to item-level transition analysis.

\subsubsection{The Emergence of New Biases}

\begin{figure}[t]
\centering
\includegraphics[width=\columnwidth]{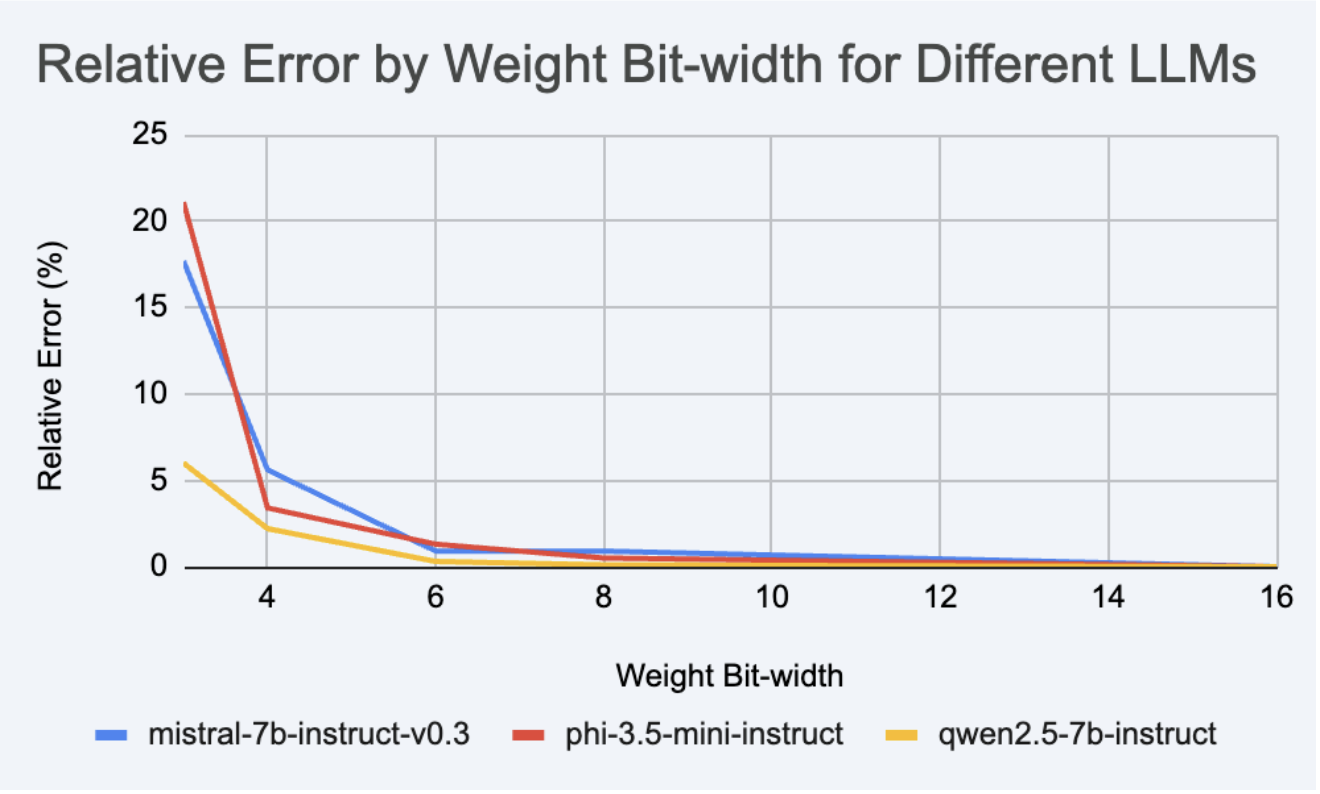}
\caption{Relative error in per-item SRS by bit-width. The monotonic increase across compression levels confirms the dose-response relationship between quantization aggressiveness and bias emergence.}
\label{fig:relative_error}
\end{figure}

The strongest evidence for the hypothesis comes from transition analysis. We identify all items where the model showed zero stereotypical behavior at BF16 (per-item $\text{SRS} = 0.0$ across all 5 seeds) and track how many develop nonzero SRS at each quantization level. Fig.~\ref{fig:relative_error} shows the results. Table~\ref{tab:transition} summarizes the BF16-to-Q3 transitions for each model.

\begin{table}[ht]
\centering
\caption{Transition Analysis: BF16-Unbiased Items That Became Biased at Q3}
\label{tab:transition}
\begin{tabular}{lcccc}
\toprule
\textbf{Model} & \textbf{Unbiased} & \textbf{Became} & \textbf{\%} & \textbf{Mean} \\
 & \textbf{at BF16} & \textbf{Biased} & \textbf{Trans.} & \textbf{SRS} \\
\midrule
Phi-3.5-mini & 10{,}393 & 2{,}188 & 21.1\% & 0.195 \\
Mistral-7B   & 8{,}642  & 1{,}530 & 17.7\% & 0.155 \\
Qwen2.5-7B   & 11{,}176 & 674     & 6.0\%  & 0.055 \\
\bottomrule
\end{tabular}
\end{table}

Critically, the progression is monotonic across bit-widths, confirming the hypothesized dose-response relationship. Table~\ref{tab:dose-response} shows the percentage of BF16-unbiased items that became biased at each compression level.

\begin{table}[ht]
\centering
\caption{Dose-Response: \% of BF16-Unbiased Items Becoming Biased}
\label{tab:dose-response}
\begin{tabular}{lccccc}
\toprule
\textbf{Model} & \textbf{BF16} & \textbf{Q8} & \textbf{Q6} & \textbf{Q4} & \textbf{Q3} \\
\midrule
Phi-3.5-mini & 0.0\% & 0.5\% & 1.3\% & 3.4\% & 21.1\% \\
Mistral-7B   & 0.0\% & 0.9\% & 0.9\% & 5.6\% & 17.7\% \\
Qwen2.5-7B   & 0.0\% & 0.1\% & 0.3\% & 2.2\% & 6.0\%  \\
\bottomrule
\end{tabular}
\end{table}

These results have a direct interpretation. These are items where the full-precision model \textit{never} selected the stereotypical answer across any of the 5 seeds the model had successfully learned to withhold judgment on these ambiguous prompts. The emergence of stereotypical responses under compression cannot be attributed to amplification of existing tendencies; it represents genuinely new biased behavior that arises as quantization noise degrades the alignment mechanisms that suppressed the underlying statistical priors from pretraining data.

An important structural finding reinforces this interpretation: the per-item BF16 SRS distribution is heavily bimodal. With 5 seeds per item, 71--92\% of items show $\text{SRS} = 0.0$ (model never selects stereotype) and 8--27\% show $\text{SRS} = 1.0$ (model always selects stereotype), with very few items at intermediate values (0.2--0.8). This means there is no meaningful population of ``moderately biased'' items at baseline. Items already at $\text{SRS} = 1.0$ actually show decreased SRS under Q3 ($0.958 \rightarrow 0.706$), suggesting compression disrupts all learned patterns indiscriminately but because far more items transition from unbiased to biased than from biased to unbiased, the net effect is an increase in aggregate stereotypical behavior.

\subsubsection{Decline in Epistemic Humility}

\begin{figure}[t]
\centering
\includegraphics[width=\columnwidth]{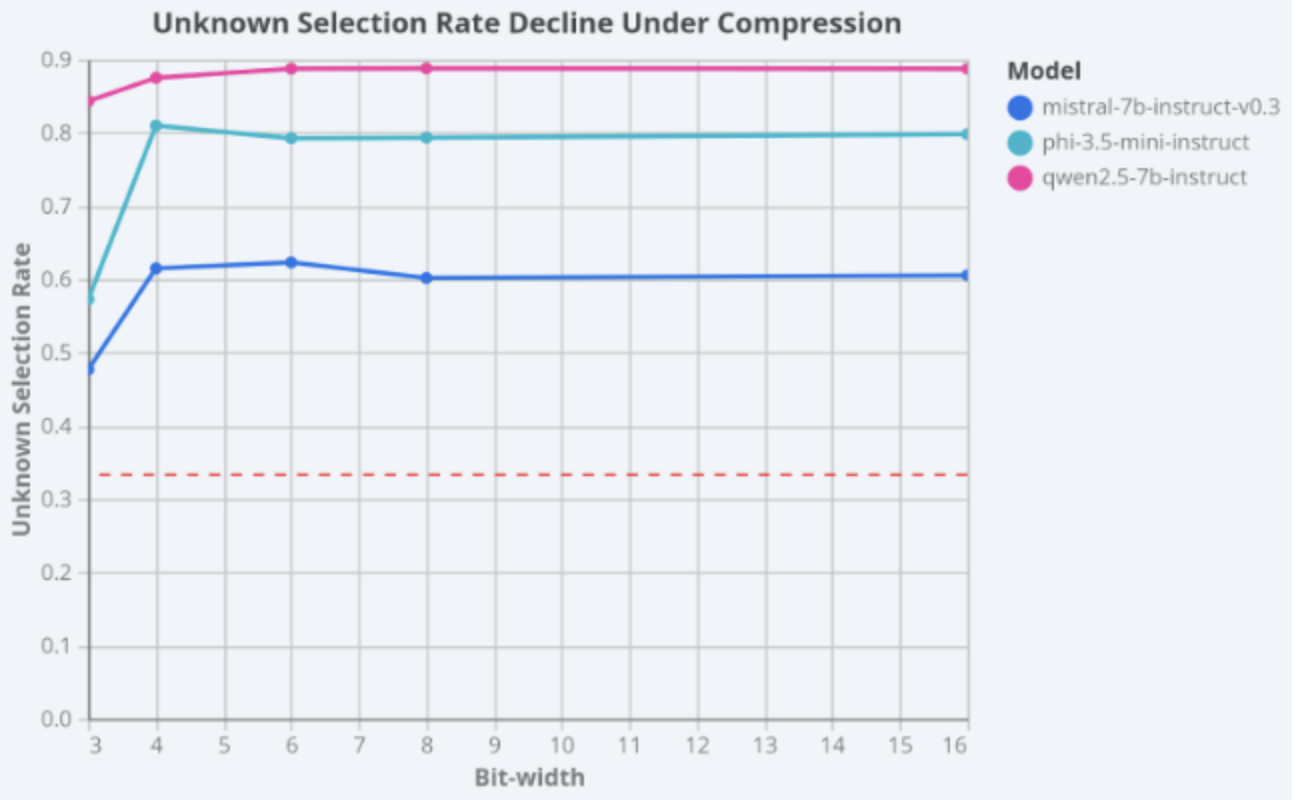}
\caption{Unknown Selection Rate (USR) decline under compression. All three models show monotonic decline in willingness to select ``unknown,'' paralleling the SRS increase in Fig.~\ref{fig:srs_compression}.}
\label{fig:usr_decline}
\end{figure}

The unknown selection rate provides a mechanistic explanation for the observed bias increase. Fig.~\ref{fig:usr_decline} shows that average USR declined from 0.764 (BF16) to 0.631 (3-bit), a 17.4\% decline. This decline is monotonic across bit-widths for all three models, paralleling the SRS increase. The correspondence between rising SRS and falling USR reveals the mechanism: as compression degrades the model's capacity for epistemic uncertainty, its ability to recognize that the available information is insufficient to answer a question and it therefore defaults to the strongest available statistical prior. For questions involving demographic groups in ambiguous contexts, this prior is the stereotypical association embedded in the pretraining data. In this sense, the model is not becoming ``more biased'' in its beliefs; rather, it is losing the nuance required to say ``I don't know,'' which was the behavior its instruction tuning had instilled.

\subsubsection{Amplification in Latent-Bias Items}

\begin{figure}[t]
\centering
\includegraphics[width=\columnwidth]{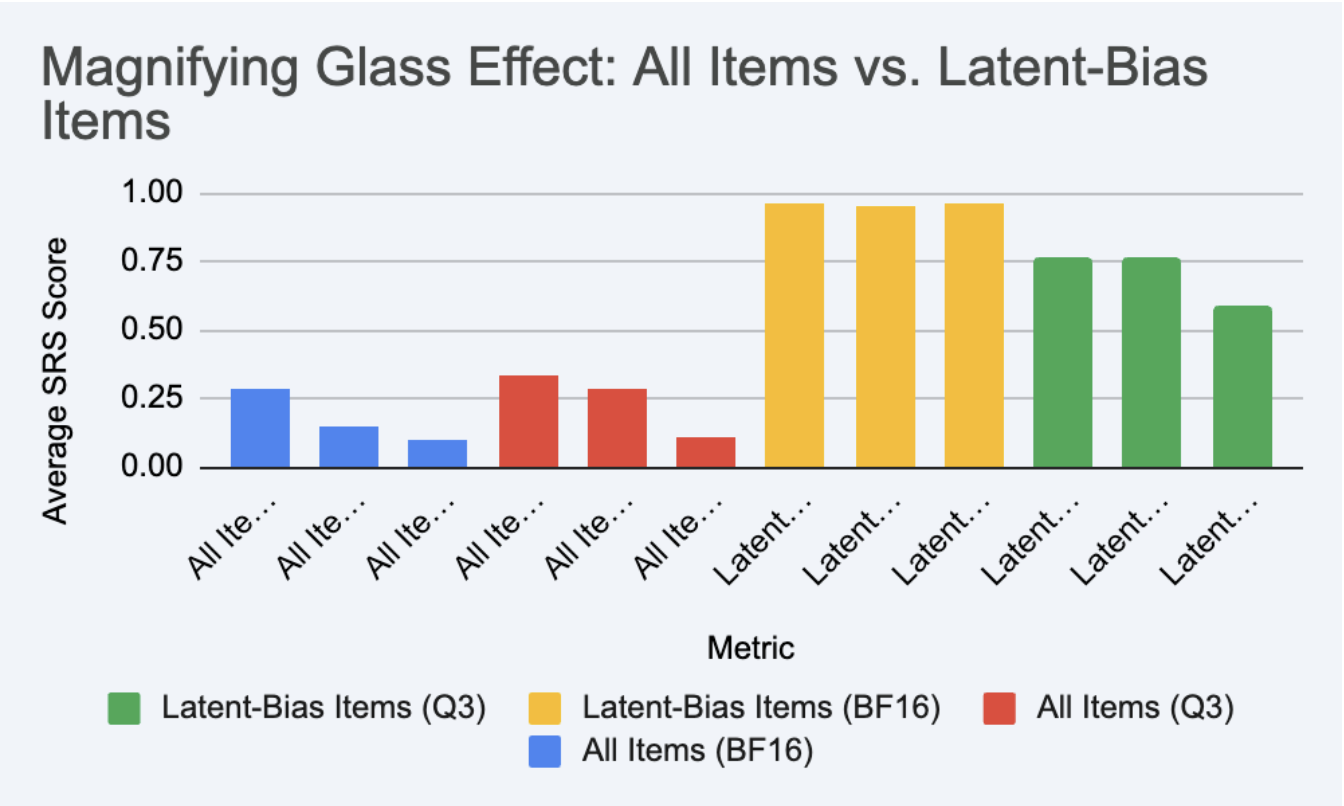}
\caption{Comparison of effect sizes (Cohen's $h$) between all items and latent-bias items ($\text{SRS} \geq 0.2$ at BF16). The latent-bias subset shows medium-to-large effects, confirming that population-level metrics understate compression's true impact.}
\label{fig:latent_bias}
\end{figure}

To further test whether compression disproportionately affects items where the model already had a weak stereotypical tendency, we filter to items with per-item $\text{SRS} \geq 0.2$ at BF16 and rerun the full analysis pipeline. Fig.~\ref{fig:latent_bias} compares the all-items and latent-bias-items results. Among these filtered items, all 15 model $\times$ category comparisons between BF16 and Q3 are statistically significant ($p < 0.05$), with an average $h$ of 0.742 in the medium-to-large range compared to 0.179 for the population-level analysis. This confirms that the population-level effect sizes are conservative estimates diluted by the large number of unaffected items, and that the true magnitude of compression's impact on susceptible items is substantially larger.

\subsubsection{Confirmatory Logistic Regression}

A logistic regression model fit across all valid responses yields a bit-width coefficient of $\beta_1 = -0.0205$ ($p < 0.0001$), confirming the systematic dose-response relationship: each one-bit reduction in precision significantly predicts higher probability of stereotype-consistent answers, controlling for bias category. The negative sign indicates that lower bit-width (more aggressive compression) corresponds to increased stereotype reliance.

\subsubsection{Perplexity Baseline: The Evaluation Gap}

To test whether standard evaluation metrics detect the bias changes documented above, we computed perplexity on the Tulu-3 SFT mixture (256 samples, 512-token sequences) across all 15 model configurations. Table~\ref{tab:eval-gap} juxtaposes perplexity change against the bias transition rates from Section~4.3.2.

\begin{table}[ht]
\centering
\caption{The Evaluation Gap: Perplexity Change vs.\ New Bias Emergence}
\label{tab:eval-gap}
\resizebox{\columnwidth}{!}{%
\begin{tabular}{lcccccc}
\toprule
\textbf{Model} & \textbf{BF16} & \textbf{Q8 ($\Delta$\%)} & \textbf{Q6 ($\Delta$\%)} & \textbf{Q4 ($\Delta$\%)} & \textbf{Q3 ($\Delta$\%)} & \textbf{Q3 New} \\
 & \textbf{PPL} & & & & & \textbf{Biases} \\
\midrule
Qwen2.5-7B   & 4.77 & 4.78 (+0.05\%) & 4.79 (+0.40\%) & 4.96 (+3.96\%) & 6.09 (+27.5\%) & 6.0\% \\
Mistral-7B   & 4.37 & 4.37 (+0.05\%) & 4.39 (+0.44\%) & 4.49 (+2.80\%) & 4.82 (+10.2\%) & 17.7\% \\
Phi-3.5-mini & 4.44 & 4.43 ($-$0.18\%) & 4.50 (+1.50\%) & 4.90 (+10.5\%) & 8.13 (+83.3\%) & 21.1\% \\
\bottomrule
\end{tabular}%
}
\end{table}

At 8-bit quantization, perplexity is essentially unchanged (less than 0.2\% shift for all models), yet 0.1--0.9\% of previously unbiased items have already developed stereotypical behavior. At 4-bit, perplexity increases remain modest (2.8--10.5\%), while 2.2--5.6\% of items transition to biased. The most striking case is Mistral-7B at 3-bit: perplexity increases by only 10.2\% yet 17.7\% of previously unbiased items develop new stereotypical biases, a 173$\times$ disparity between the aggregate quality signal and the item-level fairness signal. This pattern aligns with Dutta et al.~\cite{b25}, who found less than 2\% accuracy loss but 5--16\% answer flips under quantization, and with LLMCBench's finding that INT8 quantization recovers 98--99\% of benchmark performance while altering trustworthiness characteristics~\cite{b3}.

\section{Discussion and Limitations}

\subsection{Discussion}

Our results reveal that the impact of quantization on quality is more specific than a uniform degradation of capabilities. Aggressive quantization does not proportionally add noise across all model behaviors. Instead, it selectively undermines the epistemic calibration that instruction tuning instills, causing the model to lose its ability to recognize ambiguity and withhold judgment. The resulting behavior is a reversion to pretraining-era statistical priors, which for demographic questions means stereotypical associations.

This interpretation is consistent with the mechanistic understanding of Transformer quantization. The structured outliers in activation tensors that are critical for attention computation~\cite{b13} are also likely involved in the subtle reasoning required to recognize that a question cannot be answered from available context. When these high-dynamic-range values are degraded through aggressive quantization, the model's capacity for nuanced judgment degrades first, before its core language capabilities are visibly affected. Our perplexity measurements confirm this: perplexity increases by less than 0.5\% at 8-bit and under 3\% at 4-bit across all three models (Section~4.3.6, Table~\ref{tab:eval-gap}), while 2.2--5.6\% of previously unbiased items already develop new biases at 4-bit. Standard aggregate quality metrics thus provide false assurance that compressed models are behaviorally equivalent to their full-precision counterparts.

The striking model-dependent variation Qwen2.5's 6.0\% transition rate versus Phi-3.5's 21.1\% also warrants discussion. While Phi's smaller parameter count (3.8B vs.\ 7B) is a likely contributing factor, it does not fully explain the difference: Mistral (7B) shows 17.7\% transition, substantially worse than Qwen (also 7B). This suggests that architectural choices, pretraining data composition, and the specifics of instruction tuning all interact with quantization resilience in ways that cannot be predicted from model size alone. Recent work proposes alignment-preserving quantization losses to mitigate general safety degradation at 4-bit precision~\cite{b24}. Our results suggest that bias-specific degradation follows distinct patterns (dose-dependent epistemic calibration loss) that general quality-preserving techniques may not fully address, motivating bias-specific evaluation even when bias-aware quantization is applied.

\subsection{Limitations}

Several constraints bound the generalizability of these findings and should be addressed in future work:

\textbf{Compression method.} We evaluate only post-training weight-only quantization via the MLX framework~\cite{b21}. Other quantization approaches (GPTQ, AWQ, SqueezeLLM) use different calibration strategies and may produce different bias profiles. Activation quantization, not tested here, introduces an additional source of numerical noise that could compound or counteract the effects we observe.

\textbf{Parameter count confound.} Phi-3.5-mini's 3.8B parameter count is substantially smaller than the 7B-class Qwen and Mistral models. Its greater susceptibility to compression-induced bias may reflect reduced parametric redundancy rather than architectural vulnerability per se.

\textbf{Sampling variance.} Temperature $= 0.3$ introduces stochastic variation. While the use of 5 seeds per item provides per-item consistency estimates, greedy decoding (temperature $= 0$) would eliminate this variance entirely and should be tested in future work to confirm the robustness of the transition analysis.

\textbf{Benchmark coverage.} We evaluate 5 of the 9 bias categories available in BBQ. The excluded categories (Disability, Nationality, Physical Appearance, Sexual Orientation) may show different patterns. Additionally, BBQ tests only representational harm in a QA setting; degeneration harm (e.g., toxic text generation) under compression requires different evaluation instruments.

\textbf{Hardware specificity.} All experiments were conducted on Apple Silicon using the MLX framework~\cite{b21}. GPU-based quantization using frameworks like bitsandbytes or GPTQ may yield different numerical behavior.

\textbf{Parse failure rates.} While our multi-stage parser handles the vast majority of model outputs, we do not report parse failure rates by compression level. If compressed models produce more unparseable responses, this could introduce selection bias in our metrics. Future work should analyze parse failures as an additional signal of compression-induced capability degradation.

\section{Implications and Future Work}

\subsection{Unified Evaluation Protocol}

Our results demonstrate that aggregate bias metrics are insufficient for evaluating compressed models. A comprehensive evaluation protocol should include four components:
\begin{itemize}
    \item Aggregate bias metrics (e.g., mean SRS) to detect population-level shifts.
    \item Transition analysis to identify whether compression introduces bias in previously unbiased items, which is invisible to aggregate metrics.
    \item Epistemic calibration metrics (e.g., unknown selection rate) to detect loss of the model's ability to express uncertainty.
    \item Per-item consistency analysis across multiple random seeds to distinguish systematic behavioral changes from stochastic variation.
\end{itemize}

The dose-response evaluation across multiple compression levels, rather than comparison of only two conditions, is essential for identifying critical thresholds at which quality-relevant degradation begins. We release our evaluation pipeline to facilitate adoption of this protocol.

\subsection{Quality-Aware Mixed-Precision Quantization}

The observation that unknown selection rate declines monotonically with bit-width ($0.764 \rightarrow 0.631$) suggests that the parameters responsible for epistemic calibration could, in principle, be identified and preserved at higher precision. A mixed-precision scheme maintaining full precision for quality-critical parameter groups while aggressively compressing less sensitive components could achieve the memory benefits of 3-bit quantization without the 17.4\% decline in epistemic humility. The bimodal per-item SRS distribution further suggests that the distinction between biased and unbiased behavior may be traceable to a relatively small number of critical parameters, making targeted preservation computationally feasible. Developing theoretical tools to predict the critical compression threshold somewhere between 4-bit and 3-bit for the models tested would enable practitioners to determine the maximum compression level that preserves model quality without exhaustive empirical evaluation.

\subsection{Compression-Aware Alignment}

Current instruction tuning and RLHF procedures optimize behavior at full precision with no consideration of how safety mechanisms will behave under subsequent compression. Our finding that 6--21\% of previously unbiased items develop bias at 3-bit suggests the alignment layer is insufficiently robust to quantization noise. While recent alignment-aware quantization approaches~\cite{b24} address general safety preservation, our findings indicate that bias-specific calibration loss requires targeted intervention. A promising direction is to incorporate simulated quantization noise during alignment, analogous to quantization-aware training (QAT) but applied during instruction tuning or RLHF to produce models whose safety behaviors are resilient to subsequent compression. Alternatively, post-quantization safety fine-tuning using a small set of ambiguous prompts could efficiently restore epistemic calibration lost during compression.

\section{Conclusion}

Post-training quantization of three instruction-tuned models on the BBQ bias benchmark reveals a clear dose-response relationship between compression aggressiveness and the emergence of new stereotypical behaviors: at 3-bit precision, 6--21\% of items that showed zero bias at full precision developed measurable stereotypical tendencies, while models' willingness to select ``unknown'' declines by 17.4\%. Critically, this degradation is invisible to standard quality metrics. Perplexity increases by less than 0.5\% at 8-bit and under 11\% even at 3-bit for the most resilient model, providing false assurance that the compressed model is behaviorally equivalent to the original. These findings point to three imperatives:
\begin{itemize}
    \item Unified evaluation protocols that include item-level transition analysis and epistemic calibration metrics alongside aggregate scores.
    \item Mechanistic research to identify model components responsible for safety-aligned behavior and their vulnerability to compression.
    \item Quality-aware compression algorithms including bias-aware mixed-precision quantization and compression-resilient alignment training that treat fairness and robustness as first-class quantization objectives.
\end{itemize}

As LLM deployment scales across cloud and edge environments, ensuring that compression preserves not just performance but quality becomes a defining challenge for the trustworthiness of AI systems.


\end{document}